%% file: main.tex
\documentclass{article}

\usepackage{microtype}
\usepackage{graphicx}
\usepackage{booktabs} 

\usepackage{hyperref}

\usepackage[accepted]{icml2024}

\usepackage{amsmath}
\usepackage{amssymb}
\usepackage{mathtools}
\usepackage{amsthm}

\usepackage{booktabs}
\usepackage{adjustbox}
\usepackage{multirow}
\usepackage{algorithm}
\usepackage{amsmath}
\usepackage{subcaption}
\usepackage{diagbox}

\renewcommand{\algorithmiccomment}[1]{\textcolor{black}{\hspace*{2em} /* #1 */}}

\usepackage[capitalize,noabbrev]{cleveref}

\theoremstyle{plain}
\newtheorem{theorem}{Theorem}[section]

\newtheorem{lemma}[theorem]{Lemma}

\theoremstyle{definition}
\newtheorem{definition}[theorem]{Definition}

\theoremstyle{remark}

\usepackage[textsize=tiny]{todonotes}

\icmltitlerunning{SparseTSF: Modeling LTSF with 1k Parameters}

\begin{document}

\twocolumn[
\icmltitle{SparseTSF: Modeling Long-term Time Series Forecasting with \textit{1k} Parameters}



\icmlsetsymbol{equal}{*}

\begin{icmlauthorlist}
\icmlauthor{Shengsheng Lin}{scut}
\icmlauthor{Weiwei Lin}{scut,pcl}
\icmlauthor{Wentai Wu}{ju}
\icmlauthor{Haojun Chen}{scut}
\icmlauthor{Junjie Yang}{scut}
\end{icmlauthorlist}

\icmlaffiliation{scut}{School of Computer Science and Engineering, South China University of Technology, Guangzhou 510006, China}
\icmlaffiliation{pcl}{Peng Cheng Laboratory, Shenzhen 518066, China}
\icmlaffiliation{ju}{College of Information Science and Technology, Jinan University, Guangzhou 510632, China}

\icmlcorrespondingauthor{Weiwei Lin}{linww@scut.edu.cn}

\icmlkeywords{Machine Learning, Time Series, Long-term Time Series Forecasting, Lightweighting Technique}

\vskip 0.3in
]



\printAffiliationsAndNotice{}  

\begin{abstract}
This paper introduces SparseTSF, a novel, extremely lightweight model for Long-term Time Series Forecasting (LTSF), designed to address the challenges of modeling complex temporal dependencies over extended horizons with minimal computational resources. At the heart of SparseTSF lies the Cross-Period Sparse Forecasting technique, which simplifies the forecasting task by decoupling the periodicity and trend in time series data. This technique involves downsampling the original sequences to focus on cross-period trend prediction, effectively extracting periodic features while minimizing the model's complexity and parameter count. Based on this technique, the SparseTSF model uses fewer than \textit{1k} parameters to achieve competitive or superior performance compared to state-of-the-art models. Furthermore, SparseTSF showcases remarkable generalization capabilities, making it well-suited for scenarios with limited computational resources, small samples, or low-quality data. The code is publicly available at this repository: https://github.com/lss-1138/SparseTSF.
\end{abstract}

\section{Introduction}
\label{introduction}

Time series forecasting holds significant value in domains such as traffic flow, product sales, and energy consumption, as accurate predictions enable decision-makers to plan proactively. Achieving precise forecasts typically relies on powerful yet complex deep learning models, such as RNNs~\citep{RNN_NC}, TCNs~\citep{TCN,TCN2}, and Transformers~\citep{trans_time_survey}. In recent years, there has been a growing interest in Long-term Time Series Forecasting (LTSF), which demands models to provide an extended predictive view for advanced planning \citep{informer}.

Although a longer predictive horizon offers convenience, it also introduces greater uncertainty \citep{segrnn}. This demands models capable of extracting more extensive temporal dependencies from longer historical windows. Consequently, modeling becomes more complex to capture these long-term temporal dependencies. For instance, Transformer-based models often have millions or tens of millions of parameters, limiting their practical usability, especially in scenarios with restricted computational resources \citep{HDformer}.

In fact, the basis for accurate long-term time series forecasting lies in the inherent periodicity and trend of the data. For example, long-term forecasts of household electricity consumption are feasible due to the clear daily and weekly patterns in such data. Particularly for daily patterns, if we \textit{resample} the electricity consumption at a certain time of the day into a daily sequence, each subsequence exhibits similar or consistent trends. In this case, the original sequence's periodicity and trend are decomposed and transformed. That is, periodic patterns are transformed into \textit{inter-subsequence} dynamics, while trend patterns are reinterpreted as \textit{intra-subsequence} characteristics. This decomposition offers a novel perspective for designing lightweight LTSF models.

\begin{figure}[!htb]
    \centering
    \includegraphics[width=\linewidth]{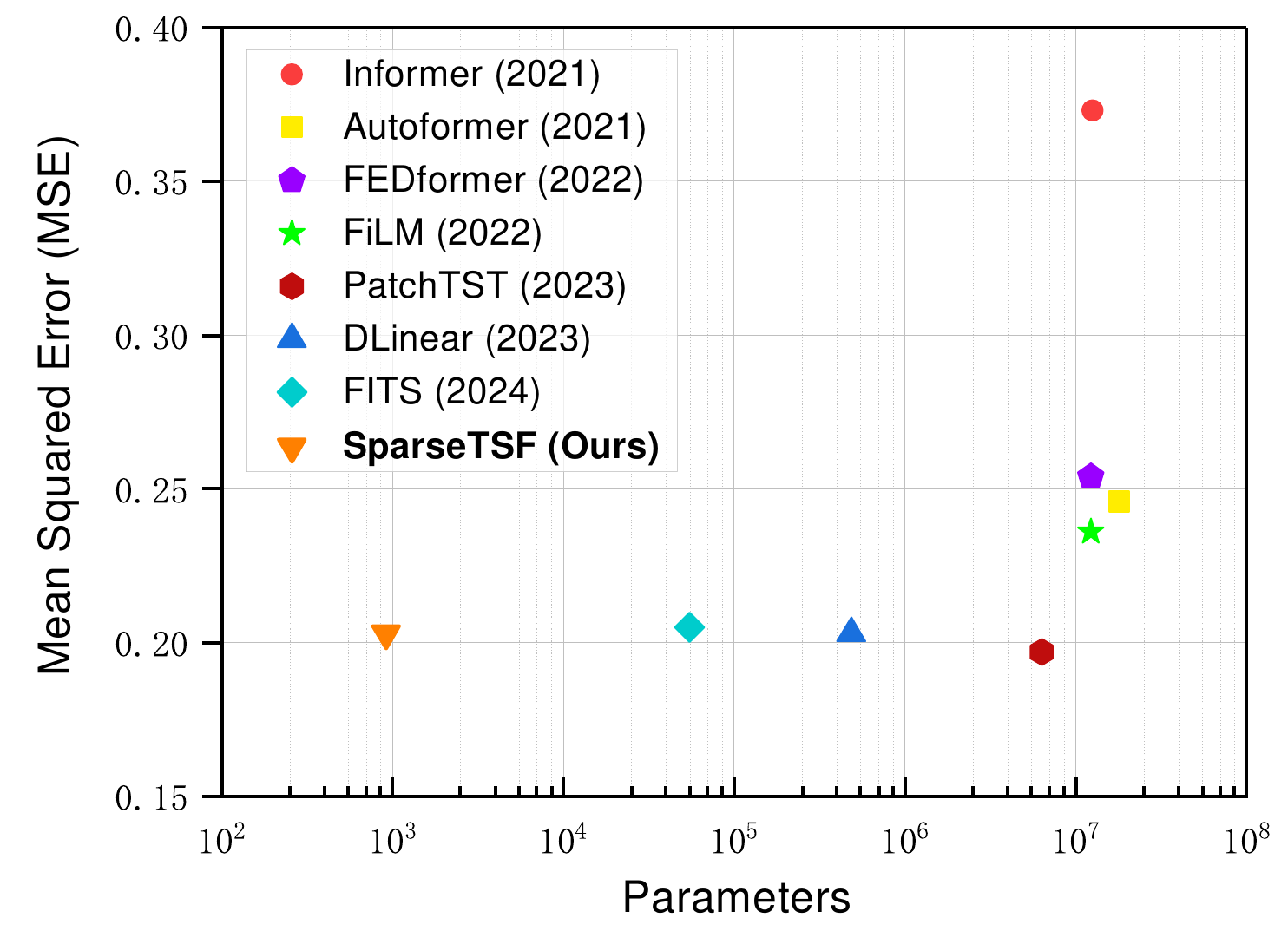}
    \caption{Comparison of MSE and parameters between SparseTSF and other mainstream models on the Electricity dataset with a forecast horizon of 720.}
    \label{fig1}
\end{figure}

In this paper, we pioneer the exploration of how to utilize this inherent periodicity and decomposition in data to construct specialized lightweight time series forecasting models. Specifically, we introduce \textbf{\textit{SparseTSF}}, an extremely lightweight LTSF model. Technically, we propose the \textit{Cross-Period Sparse Forecasting} technique (hereinafter referred to as Sparse technique). It first downsamples the original sequences with constant periodicity into subsequences, then performs predictions on each downsampled subsequence, simplifying the original time series forecasting task into a cross-period trend prediction task. This approach yields two benefits: (i) effective decoupling of data periodicity and trend, enabling the model to stably identify and extract periodic features while focusing on predicting trend changes, and (ii) extreme compression of the model's parameter size, significantly reducing the demand for computational resources. As shown in Figure~\ref{fig1}, SparseTSF achieves near state-of-the-art prediction performance with less than \textbf{\textit{1k}} trainable parameters, which makes it \textit{1\(\sim\)4 orders of magnitude smaller} than its counterparts.

In summary, our contributions in this paper are as follows:

\begin{itemize}
    \item We propose a novel \textit{Cross-Period Sparse Forecasting} technique, which downsamples the original sequences to focus on cross-period trend prediction, effectively extracting periodic features while minimizing the model’s complexity and parameter count.
    \item Based on the Sparse technique, we present the \textit{SparseTSF} model, which requires fewer than \textit{1k} parameters, significantly reducing the computational resource demand of forecasting models.
    \item The proposed SparseTSF model not only attains competitive or surpasses state-of-the-art predictive accuracy with a remarkably minimal parameter scale but also demonstrates robust generalization capabilities.
\end{itemize}

\section{Related Work}
\label{related_work}

\paragraph{Development of Long-term Time Series Forecasting}
The LTSF tasks, which aim at predicting over an extended horizon, are inherently more challenging. Initially, the Transformer architecture \citep{transformer}, known for its robust long-term dependency modeling capabilities, gained widespread attention in the LTSF domain. Models such as Informer~\citep{informer}, Autoformer~\citep{autoformer}, and FEDformer~\citep{fedformer} have modified the native structure of Transformer to suit time series forecasting tasks. More recent advancements, like PatchTST~\citep{patchtst} and PETformer~\citep{petformer}, demonstrate that the original Transformer architecture can achieve impressive results with an appropriate patch strategy, a technique that is prevalently employed in the realm of computer vision \citep{vit,MAE}. Besides Transformer architectures, Convolutional Neural Networks (CNNs) and Multilayer Perceptrons (MLPs) are also mainstream approaches, including SCINet~\citep{scinet}, TimesNet~\citep{timesnet}, MICN~\citep{micn}, TiDE~\citep{tide}, and HDMixer~\citep{HDMixer}. Recent studies have shown that transferring pretrained Large Language Models (LLMs) to the time series domain can also yield commendable results \citep{llm4ts, timellm, promptcast}. Moreover, recent works have revealed that RNN and GNN networks can also perform well in LTSF tasks, as exemplified by SegRNN~\citep{segrnn} and CrossGNN~\citep{crossgnn}.

\paragraph{Progress in Lightweight Forecasting Models}
Since DLinear~\citep{dlinear} demonstrated that simple models could already extract strong temporal periodic dependencies, numerous studies have been pushing LTSF models towards lightweight designs, including LightTS\citep{lightts}, TiDE~\citep{tide}, TSMixer~\citep{TSMixer}, and HDformer~\citep{HDformer}. Recently, FITS emerged as a milestone in the lightweight LTSF process, being the first to reduce the LTSF model scale to the \textit{10k} parameter level while maintaining excellent predictive performance \citep{fits}. FITS achieved this by transforming time-domain forecasting tasks into frequency-domain ones and using low-pass filters to reduce the required number of parameters. In this paper, our proposed SparseTSF model takes lightweight model design to the extreme. Utilizing the Cross-Period Sparse Forecasting technique, it's the first to reduce model parameters to below \textit{1k}.

\section{Methodology}
\label{methodology}
\subsection{Preliminaries}

\paragraph{Long-term Time Series Forecasting}
The task of LTSF involves predicting future values over an extended horizon using previously observed multivariate time series (MTS) data. It is formalized as \(\bar{x}_{t+1:t+H} = f(x_{t-L+1:t})\), where \(x_{t-L+1:t} \in \mathbb{R}^{L \times C}\) and \(\bar{x}_{t+1:t+H} \in \mathbb{R}^{H \times C}\). In this formulation, \(L\) represents the length of the historical observation window, \(C\) is the number of distinct features or channels, and \(H\) is the length of the forecast horizon. The main goal of LTSF is to extend the forecast horizon \(H\) as it provides richer and more advanced guidance in practical applications. However, an extended forecast horizon \(H\) also increases the complexity of the model, leading to a significant increase in parameters in mainstream models. To address this challenge, our research focuses on developing models that are not only extremely lightweight but also robust and effective.

\paragraph{Channel Independent Strategy}
Recent advancements in the field of LTSF have seen a shift towards a Channel Independent (CI) approach, especially when dealing with multivariate time series data \citep{CIorCD}. This strategy simplifies the forecasting process by focusing on individual univariate time series within the dataset. Instead of the traditional approach, which utilizes the entire multivariate historical data to predict future outcomes, the CI method finds a shared function \(f: x^{(i)}_{t-L+1:t} \in \mathbb{R}^{L} \rightarrow \bar{x}^{(i)}_{t+1:t+H} \in \mathbb{R}^{H}\) for each univariate series. This approach provides a more targeted and simplified prediction model for each channel, reducing the complexity of accounting for inter-channel relationships.

As a result, the main goal of mainstream state-of-the-art models in recent years has shifted towards effectively predict by modeling long-term dependencies, including periodicity and trends, in univariate sequences. For instance, models like DLinear achieve this by extracting dominant periodicity from univariate sequences using a single linear layer \citep{dlinear}. More advanced models, such as PatchTST~\citep{patchtst} and TiDE~\citep{tide}, employ more complex structures on single channels to extract temporal dependencies, aiming for superior predictive performance. In this paper, we adopt this CI strategy as well and focus on how to create an even more lightweight yet effective approach for capturing long-term dependencies in single-channel time series.

\subsection{SparseTSF}

\begin{figure*}[!htb]
    \centering
    \includegraphics[width=0.9\linewidth]{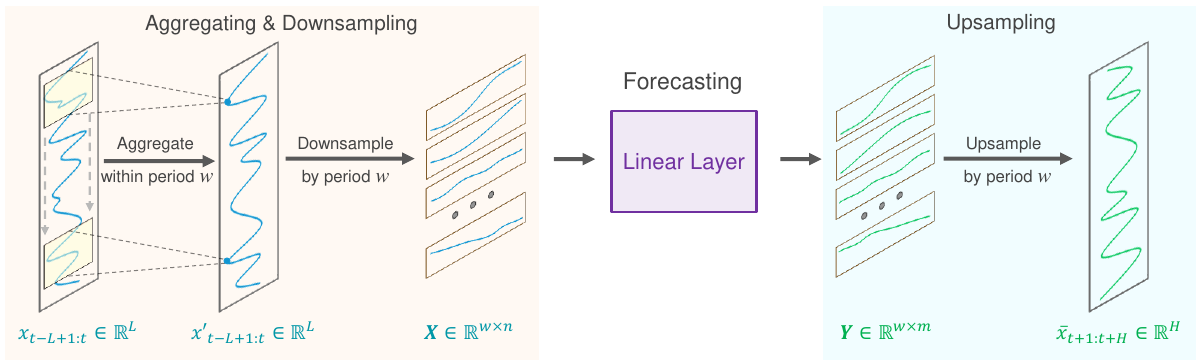}
    \caption{SparseTSF architecture.}
    \label{fig2}
\end{figure*}

Given that the data to be forecasted often exhibits constant, periodicity a priori (e.g., electricity consumption and traffic flow typically have fixed daily cycles), we propose the Cross-Period Sparse Forecasting technique to enhance the extraction of long-term sequential dependencies while reducing the model's parameter scale. Utilizing a single linear layer to model the LTSF task within this framework leads to our SparseTSF model, as illustrated in Figure~\ref{fig2}.

\paragraph{Cross-Period Sparse Forecasting}
Assuming that the time series \(x^{(i)}_{t-L+1:t}\) has a known periodicity \(w\), the first step is to downsample the original series into \(w\) subsequences of length \(n=\left \lfloor \frac{L}{w}  \right \rfloor\). A model with shared parameters is then applied to these subsequences for prediction. After prediction, the \(w\) subsequences, each of length \(m=\left \lfloor \frac{H}{w}  \right \rfloor\), are upsampled back to a complete forecast sequence of length \(H\).

Intuitively, this forecasting process appears as a sliding forecast with a sparse interval of \(w\), performed by a fully connected layer with parameter sharing within a constant period \(w\). This can be viewed as a model performing sparse sliding prediction across periods.

Technically, the downsampling process is equivalent to reshaping \(x^{(i)}_{t-L+1:t}\) into a \(n \times w\) matrix, which is then transposed to a \(w \times n\) matrix. The sparse sliding prediction is equivalent to applying a linear layer of size \(n \times m\) on the last dimension of the matrix, resulting in a \(w \times m\) matrix. The upsampling step is equivalent to transposing the \(w \times m\) matrix and reshaping it back into a complete forecast sequence of length \(H\).

However, this approach currently still faces two issues: (i) loss of information, as only one data point per period is utilized for prediction, while the rest are ignored; and (ii) amplification of the impact of outliers, as the presence of extreme values in the downsampled subsequences can directly affect the prediction.

To address these issues, we \textit{additionally} perform a sliding aggregation on the original sequence before executing sparse prediction, as depicted in Figure~\ref{fig2}. Each aggregated data point incorporates information from other points within its surrounding period, addressing issue (i). Moreover, as the aggregated value is essentially a weighted average of surrounding points, it mitigates the impact of outliers, thus resolving issue (ii). Technically, this sliding aggregation can be implemented using a 1D convolution with zero-padding and a kernel size of \( 2 \times \left \lfloor \frac{w}{2}  \right \rfloor\ + 1\). The process can be formulated as follows:
\begin{align}
x^{(i)}_{t-L+1:t} = x^{(i)}_{t-L+1:t} + \text{Conv1D}(x^{(i)}_{t-L+1:t})
\end{align}

\paragraph{Instance Normalization}
Time series data often exhibit distributional shifts between training and testing datasets. Recent studies have shown that employing simple sample normalization strategies between the input and output of models can help mitigate this issue \citep{revin, dlinear}. In our work, we also utilize a straightforward normalization strategy. Specifically, we subtract the mean of the sequence from itself before it enters the model and add it back after the model's output. This process is formulated as follows:
\begin{align}
x^{(i)}_{t-L+1:t} = x^{(i)}_{t-L+1:t} - \mathbb{E}_t(x^{(i)}_{t-L+1:t}),\\
\bar{x}^{(i)}_{t+1:t+H} = \bar{x}^{(i)}_{t+1:t+H} + \mathbb{E}_t(x^{(i)}_{t-L+1:t}).
\end{align}

\paragraph{Loss Function}
In alignment with current mainstream practices in the field, we adopt the classic Mean Squared Error (MSE) as the loss function for SparseTSF. This function measures the discrepancy between the predicted values \(\bar{x}^{(i)}_{t+1:t+H}\) and the actual ground truth \(y^{(i)}_{t+1:t+H}\). It is formulated as:
\begin{align}
\mathcal{L} = \frac{1}{C}\sum_{i=1}^{C}\left \| y^{(i)}_{t+1:t+H} - \bar{x}^{(i)}_{t+1:t+H}  \right \|_{2}^{2}.
\end{align}

\subsection{Theoretical Analysis}
\label{theoretical_analysis}

In this section, we provide a theoretical analysis of the SparseTSF model, focusing on its parameter efficiency and the effectiveness of the Sparse technique. The relevant theoretical proofs are provided in Appendix~\ref{proofs}.

\subsubsection{Parameter Efficiency of SparseTSF}

\begin{theorem}
\label{theorem1}
Given a historical look-back window length \(L\), a forecast horizon \(H\), and a constant periodicity \(w\), the total number of parameters required for the SparseTSF model is \( \left \lfloor \frac{L}{w} \right \rfloor \times \left \lfloor \frac{H}{w} \right \rfloor + 2 \times \left \lfloor \frac{w}{2}  \right \rfloor\ + 1 \).
\end{theorem}

In LTSF tasks, the look-back window length \(L\) and forecast horizon \(H\) are usually quite large, for instance, up to 720, while the intrinsic periodicity \(w\) of the data is also typically large, such as 24. In this scenario, \( \left \lfloor \frac{L}{w} \right \rfloor \times \left \lfloor \frac{H}{w} \right \rfloor + 2 \times \left \lfloor \frac{w}{2} \right \rfloor + 1 \ll L \times H \). This means that the parameter scale of the SparseTSF model is much lighter than even the simplest single-layer linear model. This demonstrates the lightweight architecture of the SparseTSF model.

\subsubsection{Effectiveness of SparseTSF}
The time series targeted for long-term forecasting often exhibits constant periodicity. Here, we first define the representation of such a sequence \( X \).

\begin{definition}
\label{definition}
Consider a \textit{univariate} time series \( X \) with a known period \( w \), which can be decomposed into a periodic component \( P(t) \) and a trend component \( T(t) \), such that \( X(t) = P(t) + T(t) \). Here, \( P(t) \) represents the periodic part and satisfies the condition:
\begin{align}
\label{periodic_func}
P(t) = P(t + w).
\end{align}
\end{definition}

Furthermore, we can derive the form of the modeling task after downsampling.

In the context of a truncated subsequence \( x_{t-L+1:t} \) of \( X(t) \) and its corresponding future sequence \( x_{t+1:t+H} \) to be forecasted, the conventional approach involves using \( x_{t-L+1:t} \) directly to predict \( x_{t+1:t+H} \), essentially estimating the function:
\begin{align}
x_{t+1:t+H} = f(x_{t-L+1:t})
\end{align}
However, with the application of the Sparse technique, this forecasting task transforms into predicting downsampled subsequences, as per Lemma~\ref{lemma}.

\begin{lemma}
\label{lemma}
The SparseTSF model reformulates the forecasting task into predicting downsampled subsequences, namely:
\begin{align}
x'_{t+1:t+m} = f(x'_{t-n+1:t})
\end{align}
\end{lemma}

Combining Definition~\ref{definition} and Lemma~\ref{lemma}, we can further deduce Theorem~\ref{theorem2}.
\begin{theorem}
\label{theorem2}
Given a time series dataset that satisfies Definition~\ref{definition}, the SparseTSF model's formulation becomes:
\begin{align}
p'_{t+1:t+m}+t'_{t+1:t+m} = f(p'_{t-n+1:t}+t'_{t-n+1:t})
\end{align}
where, for any \(i \in [t-n+1:t+m]\) and \(j \in [t-n+1:t+m]\), satisfying:
\begin{align}
p'_i = p'_j
\end{align}
\end{theorem}

Theorem~\ref{theorem2} implies that the task of the SparseTSF model effectively transforms into predicting future \textit{trend} components (i.e., \(t'\)), using the \textit{constant} periodic components (i.e., \(p'\)) as a reference. This process effectively separates the periodic components, which are no longer explicitly modeled, allowing the model to focus more on the trend variations.

Intuitively, We can further validate this finding from the perspective of \textit{autocorrelation}, a powerful tool for identifying patterns such as seasonality or periodicity in time series data.

\begin{definition}[AutoCorrelation Function (ACF) \citep{acf}]
\label{definition2}
Given a time series \( \{X_t\} \), where \( t \) represents discrete time points, the ACF at lag \( k \) is defined as:
\begin{equation}
\text{ACF}(k) = \frac{\sum_{t=1}^{N-k} (X_t - \mu)(X_{t+k} - \mu)}{\sum_{t=1}^{N} (X_t - \mu)^2}
\end{equation}
where \( N \) is the total number of observations in the time series, \( X_t \) is the value of the series at time \( t \), \( X_{t+k} \) is the value of the series at time \( t+k \), and \( \mu \) is the mean of the series \( \{X_t\} \). 
\end{definition}

\begin{figure}[!htb]  
    \centering
    \subcaptionbox{Original}{\includegraphics[width=0.48\linewidth]{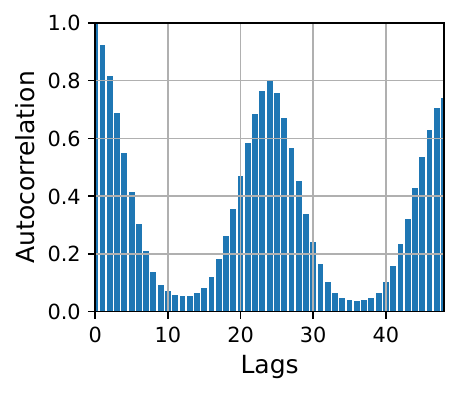}} \label{fig3a}
    \subcaptionbox{Downsampled}{\includegraphics[width=0.48\linewidth]{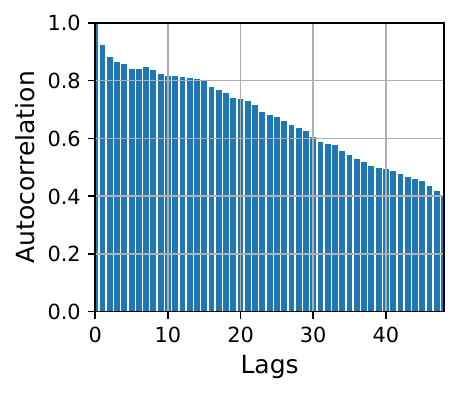}}
    \label{fig3b}
    \caption{Comparison of autocorrelation in original and downsampled subsequences for the first channel in the ETTh1 dataset.}
    \label{fig3}
\end{figure}

The lag time \( k \) in the ACF reveals the periodic patterns in the series, that is, when \( k \) equals the periodic length of the series, the ACF value typically shows a significant peak. As shown in Figure~\ref{fig3}, the original sequence exhibits clear periodicity, while the downsampled subsequences retain only trend characteristics. This demonstrates that, through its downsampling strategy, the SparseTSF model can efficiently separate and extract accurate periodic features from time series data. This not only reduces the complexity of the model but also enables it to focus on predicting trend variations, thereby exhibiting impressive performance in LTSF tasks.

In summary, the SparseTSF model's design, characterized by its parameter efficiency and focus on decoupling periodic features, makes it well-suited for LTSF tasks, especially in scenarios where the data exhibits clear periodic patterns.

\section{Experiments}
\label{experiments}

In this section, we present the experimental results of SparseTSF on mainstream LTSF benchmarks. Additionally, we discuss the efficiency advantages brought by the lightweight architecture of SparseTSF. Furthermore, we conduct ablation studies and analysis to further reveal the effectiveness of the Sparse technique.

\subsection{Experimental Setup}

\paragraph{Datasets} We conducted experiments on four mainstream LTSF datasets that exhibit daily periodicity. These datasets include ETTh1\&ETTh2\footnote{https://github.com/zhouhaoyi/ETDataset}, Electricity\footnote{https://archive.ics.uci.edu/ml/datasets}, and Traffic\footnote{https://pems.dot.ca.gov/}. The details of these datasets are presented in Table~\ref{dataset}.
\input{tables/dataset}

\paragraph{Baselines} We compared our approach with state-of-the-art and representative methods in the field. These include Informer~\citep{informer}, Autoformer~\citep{autoformer}, Pyraformer~\citep{pyraformer}, FEDformer~\citep{fedformer}, Film~\citep{film}, TimesNet~\citep{timesnet}, and PatchTST~\citep{patchtst}. Additionally, we specifically compared SparseTSF with lightweight models, namely DLinear~\citep{dlinear} and FITS~\citep{fits}. Following FITS, SparseTSF defaults to a look-back length of 720.

\paragraph{Environment} All experiments in this study were implemented using PyTorch~\citep{pytorch} and conducted on a single NVIDIA RTX 4090 GPU with 24GB of memory. More experimental details are provided in Appendix~\ref{more_detail}.

\subsection{Main Results}
\input{tables/main_result}

Table~\ref{main_reslut} presents a performance comparison between SparseTSF and other baseline models\footnote{Recent works discovered a long-standing bug in the current benchmark framework, which may affect model performance on small datasets \citep{fits,TFB}. We reporte the comparison results after fixing this bug in Appendix~\ref{fix_result}.}. It is observable that SparseTSF ranks within the top two in all scenarios, achieving or closely approaching state-of-the-art levels with a significantly smaller parameter scale. This emphatically demonstrates the superiority of the Sparse technique proposed in this paper. Specifically, the Sparse technique is capable of more effectively extracting the periodicity and trends from data, thereby enabling exceptional predictive performance in long horizon scenarios. Additionally, the standard deviation of SparseTSF's results is notably small. In most cases, the standard deviation across 5 runs is within 0.001, which strongly indicates the robustness of the SparseTSF model.

\subsection{Efficiency Advantages of SparseTSF}

Beyond its powerful predictive performance, another significant benefit of the SparseTSF model is its extreme lightweight nature. Previously, Figure~\ref{fig1} visualized the parameter-performance comparison of SparseTSF with other mainstream models. Here, we further present a comprehensive comparison between SparseTSF and these baseline models in terms of both static and runtime metrics, including:

\begin{enumerate}
    \item \textbf{Parameters}: The total number of trainable parameters in the model, representing the model's size.
    \item \textbf{MACs} (Multiply-Accumulate Operations): A common measure of computational complexity in neural networks, indicating the number of multiply-accumulate operations required by the model.
    \item \textbf{Max Memory}: The maximum memory usage during the model training process.
    \item \textbf{Epoch Time}: The training duration for a single epoch. This metric was averaged over 3 runs.
\end{enumerate}

\input{tables/measure}

Table~\ref{measure} displays the comparative results. It is evident that SparseTSF significantly outperforms other models in terms of static metrics like the number of parameters and MACs, being over ten times smaller than the next best model. This characteristic allows SparseTSF to be deployed on devices with very limited computational resources. Furthermore, in terms of runtime metrics, Max Memory and Epoch Time, SparseTSF significantly outperforms other mainstream models, rivaling the existing lightweight models (i.e., DLinear and FITS). Herein, DLinear benefits from a shorter look-back length, achieving the lowest overhead, while FITS and SparseTSF incur additional overhead due to extra operations (i.e., Fourier transformation and resampling).

\input{tables/parameters}

Additionally, we conducted a comprehensive comparison with FITS, a recent milestone work in the field of LTSF model lightweight progression. The results in Table~\ref{parameters} reveal that SparseTSF significantly surpasses FITS in terms of parameter scale under any input-output length configuration. Therefore, SparseTSF marks another significant advancement in the journey towards lightweight LTSF models.

\subsection{Ablation Studies and Analysis}
Beyond its ultra-lightweight characteristics, the Sparse technique also possesses a robust capability to extract periodic features, which we will delve further into in this section.

\input{tables/ablation}

\paragraph{Effectiveness of the Sparse Technique}
The Sparse technique, combined with a simple single-layer linear model, forms the core of our proposed model, SparseTSF. Additionally, the Sparse technique can be integrated with other foundational models, including the Transformer~\citep{transformer} and GRU~\citep{gru} models. As demonstrated in the results of Table~\ref{ablation}, the incorporation of the Sparse technique significantly enhances the performance of all models, including Linear, Transformer, and GRU. Specifically, the Linear model showed an average improvement of 4.7\%, the Transformer by 21.4\%, and the GRU by 12.4\%. These results emphatically illustrate the efficacy of the Sparse technique. Therefore, the Sparse technique can substantially improve the performance of base models in LTSF tasks.

\paragraph{Representation Learning of the Sparse Technique}
In Section~\ref{theoretical_analysis}, we theoretically analyzed the reasons why the Sparse technique can enhance the performance of forecasting tasks. Here, we further reveal the role of the Sparse technique from a representation learning perspective. Figure~\ref{fig3} shows the distribution of normalized weights for both the trained Linear model and the SparseTSF model. The weight of the Linear model is an \(L \times H\) matrix, which can be directly obtained. However, as the SparseTSF model is a sparse model, we need to acquire its equivalent weights. To do this, we first input \(H\) one-hot encoded vectors of length \(L\) into the SparseTSF model (when \(L\) equals \(H\), this can be simplified to a diagonal matrix, i.e., diagonal elements are 1, and other elements are 0). We then obtain and transpose the corresponding output to get the equivalent \(L \times H\) weight matrix of SparseTSF. When \(L\) equals \(H\), this process is formulated as:
\begin{align}
     {weight}'=\textit{SparseTSF}(\begin{bmatrix}
     1 & 0 & \dots  & 0\\
     0 & 1 & \dots & 0\\
     \dots & \dots & \dots & 0\\
     0 & 0 & 0 & 1\\
    \end{bmatrix}) ^\top.
\end{align}
\begin{figure*}[!htb]  
    \centering
    \subcaptionbox{Linear}{\includegraphics[width=0.4\linewidth]{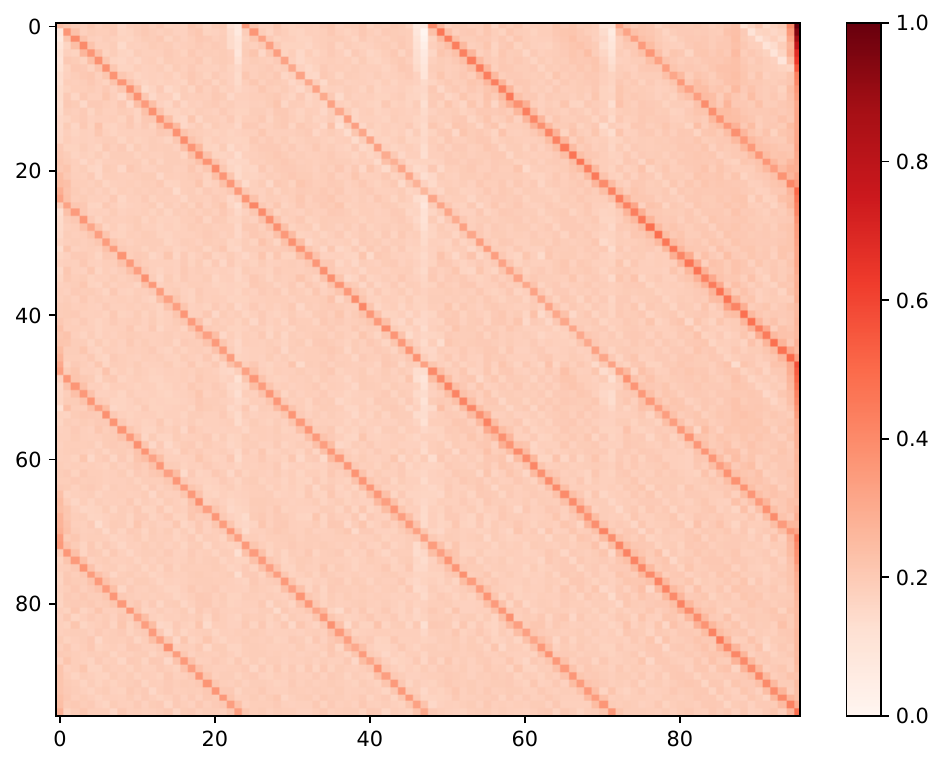}} \label{fig4a}
    \subcaptionbox{SparseTSF}{\includegraphics[width=0.4\linewidth]{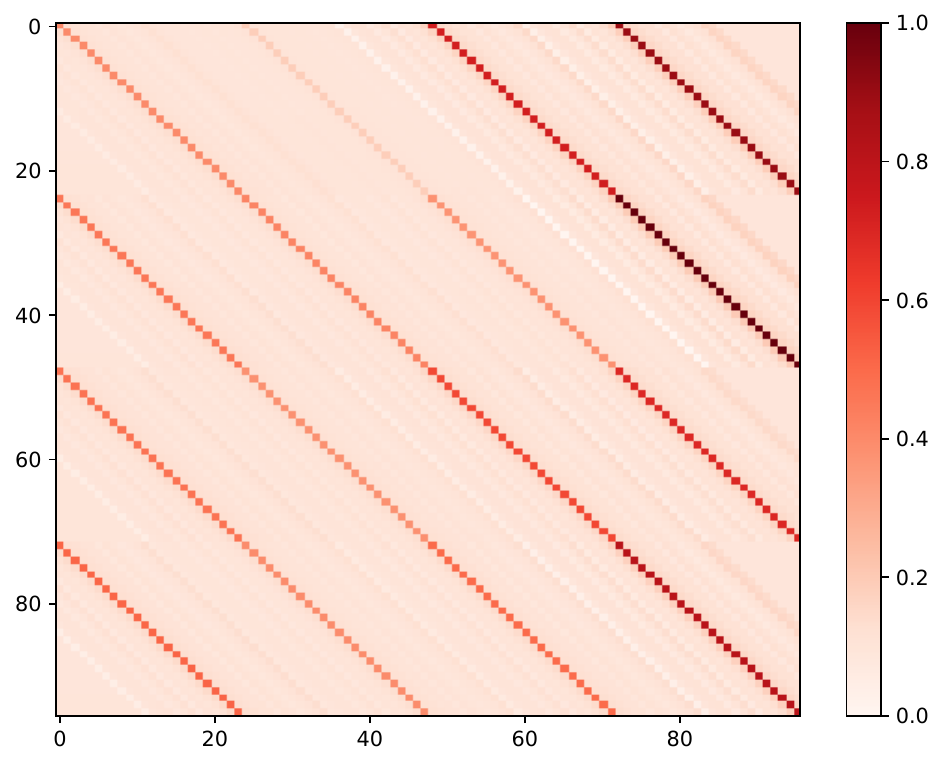}}
    \label{fig4b}
    \caption{Visualization of normalized weights of the model trained on the ETTh1 dataset with both look-back length (X-axis) and forecast horizon (Y-axis) of 96.}
    \label{fig4}
\end{figure*}

From the visualization in Figure~\ref{fig4}, two observations can be made: (i) The Linear model can learn evenly spaced weight distribution stripes (i.e., periodic features) from the data, indicating that single linear layer can already extract the primary periodic characteristics from a univariate series with the CI strategy. These findings are consistent with previous research conclusions \citep{dlinear}. (ii) Compared to the Linear model, SparseTSF learns more distinct evenly spaced weight distribution stripes, indicating that SparseTSF has a stronger capability in extracting periodic features. This phenomenon aligns with the conclusions of Section~\ref{theoretical_analysis}.

Therefore, the Sparse technique can enhance the model's performance in LTSF tasks by strengthening its ability to extract periodic features from data.

\paragraph{Impact of the Hyperparameter \(w\)}

The Sparse technique relies on the manual setting of the hyperparameter \(w\), which represents the a priori main period. Here, we delve into the influence of different values of \(w\) on the forecast outcomes. As indicated in the results from Table~\ref{period}, SparseTSF exhibits optimal performance when \(w=24\), aligning with the intrinsic main period of the data. Conversely, when \(w\) diverges from 24, a slight decline in performance is observed. This suggests that the hyperparameter \(w\) should ideally be set consistent with the data's a priori main period.

\input{tables/period}

In practical scenarios, datasets requiring long-term forecasting often exhibit inherent periodicity, such as daily or weekly cycles, common in domains like electricity, transportation, energy, and consumer goods consumption.  Therefore, empirically identifying the predominant period and setting the appropriate \(w\) for such data is both feasible and straightforward. However, for data lacking clear periodicity and patterns, such as financial data, current LTSF models may not be effective \citep{dlinear}. Thus, the SparseTSF model may not be the preferred choice for these types of data. Nonetheless, we will further discuss the existing limitations and potential improvements of the SparseTSF model in the Section~\ref{limits}.

\paragraph{Generalization Ability of the SparseTSF Model}
The Sparse technique enhances the model's ability to extract periodic features from data. Therefore, the generalization capability of a trained SparseTSF model on different datasets with the \textit{same principal periodicity} is promising. To investigate this, we further studied the cross-domain generalization performance of the SparseTSF model (i.e., training on a dataset from one domain and testing on a dataset from another). Specifically, we examined the performance from ETTh2 to ETTh1, which are datasets of the same type but collected from different machines, each with 7 variables. Additionally, we explored the performance from Electricity to ETTh1, where these datasets originate from different domains and have a differing number of variables (i.e., Electricity has 321 variables). On datasets with different numbers of variables, models trained with traditional non-CI strategies (like Informer) cannot transfer, whereas those trained with CI strategies (like PatchTST) can, due to the decoupling of CI strategies from channel relationships. These datasets all have a daily periodicity, i.e., a prior predominant period of \(w=24\).

\input{tables/cross_domain}

Experimental results, as shown in Table~\ref{cross_domain}, reveal that SparseTSF outperforms other models in both similar domain generalization (ETTh2 to ETTh1) and less similar domain generalization (Electricity to ETTh1). It is expected that performance on ETTh2 to ETTh1 would be superior to Electricity to ETTh1. Furthermore, in both scenarios, the generalization performance of SparseTSF is nearly on par with the performance of direct modeling in the SparseTSF source domain as shown in Table~\ref{main_reslut} and surpasses other baselines that model directly in the source domain. This robustly demonstrates the generalization capability of SparseTSF, indirectly proving the Sparse technique's ability to extract stable periodic features.

Therefore, the SparseTSF model exhibits outstanding generalization capabilities. This characteristic is highly beneficial for the application of the SparseTSF model in scenarios involving small samples and low-quality data.

\section{Discussion}

\subsection{Limitations and Future Work}
\label{limits}

The SparseTSF model proposed in this paper excels in handling data with a stable main period, demonstrating enhanced feature extraction capabilities and an extremely lightweight architecture. However, there are two scenarios where SparseTSF may not be as effective:
\begin{enumerate}
\item \textbf{Ultra-Long Periods}: In cases involving ultra-long periods (for example, periods exceeding 100), the Sparse technique results in overly sparse parameter connections. Consequently, SparseTSF does not perform optimally in such scenarios.
\item \textbf{Multiple Periods}: SparseTSF may struggle with data that intertwines multiple periods, as the Sparse technique can only downsample and decompose one main period.
\end{enumerate}

We have further investigated the performance of SparseTSF in these scenarios in Appendix~\ref{case_study} and concluded that: (1) in ultra-long period scenarios, a denser connected model would be a better choice; (2) SparseTSF can still perform excellently in some multi-period scenarios (such as daily periods superimposed with weekly periods).

Finally, one of our key future research directions is to further address the these potential limitations by designing additional modules to enhance SparseTSF's ability, thus achieving a balance between performance and parameter size.

\subsection{Differences Compared to Existing Methods}
The Sparse technique proposed in this paper involves downsampling/upsampling to achieve periodicity/trend decoupling. It may share a similar idea with existing methods, as downsampling/upsampling and periodic/trend decomposition techniques are prevalent in related literature nowadays. Specifically, we provide a detailed analysis of the differences with respect to N-HiTS~\citep{Nhits} and OneShotSTL~\citep{OneShotSTL} as follows, and present the comparison results in Appendix~\ref{comp_other}.

\paragraph{SparseTSF Compared to N-HiTS}
N-HiTS incorporates novel hierarchical interpolation and multi-rate data sampling techniques to achieve better results \citep{Nhits}. The downsampling and upsampling techniques proposed in SparseTSF are indeed quite different from those used in N-HiTS, including:
\begin{itemize}
    \item The downsampling and upsampling in SparseTSF occur before and after the model's prediction process, respectively, whereas N-HiTS conducts these operations within internally stacked modules.
    \item SparseTSF's downsampling involves resampling by a factor of \(w\) to \(w\) subsequences of length \(L/w\), which is technically equivalent to matrix reshaping and transposition, whereas N-HiTS employs downsampling through max-pooling.
    \item SparseTSF's upsampling involves transposing and reshaping the predicted subsequences back to the original sequence, whereas N-HiTS achieves upsampling through interpolation.
\end{itemize}

\paragraph{SparseTSF Compared to OneShotSTL}
Seasonal-trend decomposition (STD) is a classical and powerful tool for time series forecasting, and OneShotSTL makes a great contribution to advancing the lightweight long-term forecasting process, featuring fast, lightweight, and powerful capabilities \citep{OneShotSTL}. However, SparseTSF differs significantly from OneShotSTL in several aspects:
\begin{itemize}
    \item SparseTSF is a neural network model while OneShotSTL is a non-neural network method focused on online forecasting.
    \item OneShotSTL minimizes residuals and calculates trend and seasonal subseries separately from the original sequence with lengths of \(L\), whereas our SparseTSF resamples the original sequence into \(w\) subseries of length \(L/w\) with a constant period \(w\).
    \item OneShotSTL accelerates inference by optimizing the original computation for online processing, while SparseTSF achieves lightweighting by using parameter-sharing linear layers for prediction across all subseries.
\end{itemize}

\section{Conclusion}
\label{conclusion}
In this paper, we introduce the Cross-Period Sparse Forecasting technique and the corresponding SparseTSF model. Through detailed theoretical analysis and experimental validation, we demonstrated the lightweight nature of the SparseTSF model and its capability to extract periodic features effectively. Achieving competitive or even surpassing the performance of current state-of-the-art models with a minimal parameter scale, SparseTSF emerges as a strong contender for deployment in computation resource-constrained environments. Additionally, SparseTSF exhibits potent generalization capabilities, opening new possibilities for applications in transferring to small samples and low-quality data scenarios. SparseTSF stands as another milestone in the journey towards lightweight models in the field of long-term time series forecasting. Finally, we aim to further tackle the challenges associated with extracting features from ultra-long-periodic and multi-periodic data in the future, striving to achieve an optimal balance between model performance and parameter size.

\section*{Acknowledgements}
This work is supported by Guangdong Major Project of Basic and Applied Basic Research (2019B030302002), National Natural Science Foundation of China (62072187), Guangzhou Development Zone Science and Technology Project (2021GH10) and the Major Key Project of PCL, China under Grant PCL2023A09.

\section*{Impact Statement}
This paper presents work whose goal is to advance the field of Machine Learning. There are many potential societal consequences of our work, none which we feel must be specifically highlighted here.

\bibliography{myref}
\bibliographystyle{icml2024}

\newpage

\appendix
\section{More Details of SparseTSF}

\subsection{Overall Workflow}
\label{pseudocode}

The complete workflow of SparseTSF is outlined in Algorithm~\ref{alg1}, which takes a \textit{univariate} historical look-back window \(x_{t-L+1:t}\) as input and outputs the corresponding forecast results \(\bar{x}_{t+1:t+H}\). By integrating the CI strategy, i.e., modeling multiple channels using a model with shared parameters, multivariate time series forecasting can be effectively achieved.

\begin{algorithm}
\caption{The Overall Pseudocode of SparseTSF}
\label{alg1}
\begin{algorithmic}[1]
\REQUIRE Historical look-back window $x_{t-L+1:t} \in \mathbb{R}^L$
\ENSURE Forecasting horizon $\bar{x}_{t+1:t+H} \in \mathbb{R}^H$

\STATE $e_t \gets \frac{{\sum_{i=t-L+1}^{t} x_i}}{L}$ \COMMENT{Calculate the mean of the look-back window}
\STATE $x_{t-L+1:t} \gets x_{t-L+1:t} - e_t$ \COMMENT{Subtract the mean from each element}
\STATE $x_{t-L+1:t} \gets \textit{Conv1d}(x_{t-L+1:t}, 2 \times \left \lfloor \frac{w}{2}  \right \rfloor + 1) + x_{t-L+1:t}$ \COMMENT{Apply 1D convolution on the original window}
\STATE $X \gets \textit{Reshape}(x_{t-L+1:t}, (n, w))$ \COMMENT{Reshape $x_{t-L+1:t}$ into a $n \times w$ matrix}
\STATE $Y \gets \textit{Linear}(X^\top)^\top$ \COMMENT{Transpose, apply linear transformation $n \to m$, and transpose back}
\STATE $\bar{x}_{t+1:t+H} \gets \textit{Reshape}(Y, (H))$ \COMMENT{Reshape $Y$ back into a length $H$ sequence}
\STATE $\bar{x}_{t+1:t+H} \gets \bar{x}_{t+1:t+H} + e_t$ \COMMENT{Add the mean back to each element}
\end{algorithmic}
\end{algorithm}

\begin{figure}[!htb]
    \centering
    \includegraphics[width=\linewidth]{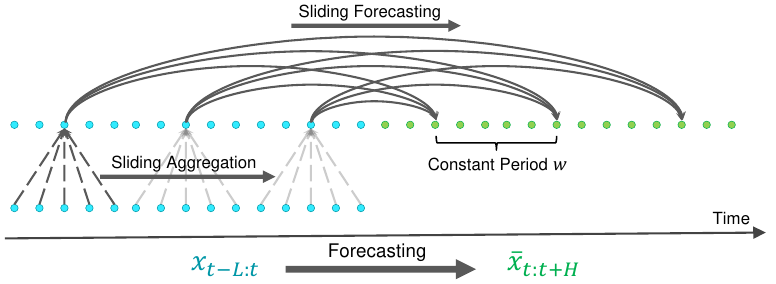}
    \caption{Schematic illustration of SparseTSF.}
    \label{fig_append}
\end{figure}

Additionally, intuitively, SparseTSF can be perceived as a sparsely connected linear layer performing sliding prediction across periods, as depicted in Figure~\ref{fig_append}.

\subsection{Experimental Details}
\label{more_detail}

We implemented SparseTSF in PyTorch~\citep{pytorch} and trained it using the Adam optimizer~\citep{adam} for 30 epochs, with a learning rate decay of 0.8 after the initial 3 epochs, and early stopping with a patience of 5. The dataset splitting follows the procedures of FITS and Autoformer, where the ETT datasets are divided into proportions of 6:2:2, while the other datasets are split into proportions of 7:1:2.

SparseTSF has minimal hyperparameters due to its simple design. The period \(w\) is set to the inherent cycle of the data (e.g., \(w=24\) for ETTh1) or to a smaller value if the data has an extremely long cycle (e.g., \(w=4\) for ETTm1). The choice of batch size depends on the size of the data samples (i.e., the number of channels). For datasets with fewer than 100 channels (such as ETTh1), the batch size is set to 256, while for datasets with fewer than 300 channels (such as Electricity), the batch size is set to 128. This setting maximizes the utilization of GPU parallel computing capabilities while avoiding GPU out-of-memory issues (i.e., with NVIDIA RTX 4090, 24GB). Additionally, the learning rate needs to be set relatively large (i.e., 0.02) due to the very small number of learnable parameters in SparseTSF. The complete details can be found in our official repository\footnote{https://github.com/lss-1138/SparseTSF}.

The baseline results in this paper are from the first version of the FITS paper\footnote{https://arxiv.org/pdf/2307.03756v1.pdf}, where FITS adopted a uniform input length of 720 (we also use an input length of 720 for fair comparison with it). Here, the input lengths of other baselines are set to be consistent with their respective official input lengths.

\section{Theoretical Proofs}
\label{proofs}
\paragraph{Proof of Theorem~\ref{theorem1}}
\begin{proof}
The SparseTSF model consists of two main components: a 1D convolutional layer for sliding aggregation and a linear layer for sparse sliding prediction. The number of parameters in the 1D convolutional layer (without bias) is determined by the kernel size, which is \(2 \times \left \lfloor \frac{w}{2}  \right \rfloor\ + 1\). For the linear layer (without bias), the number of parameters is the product of the input and output sizes, which are \(n = \left \lfloor \frac{L}{w} \right \rfloor\) and \(m = \left \lfloor \frac{H}{w} \right \rfloor\), respectively. Thus, the total number of parameters in the linear layer is \(n \times m\).

By combining the parameters from both layers, the total count is:
\(
 n \times m + 2 \times \left \lfloor \frac{w}{2}  \right \rfloor\ + 1= \left \lfloor \frac{L}{w} \right \rfloor \times \left \lfloor \frac{H}{w} \right \rfloor + 2 \times \left \lfloor \frac{w}{2}  \right \rfloor\ + 1.
\)
\end{proof}

\paragraph{Proof of Lemma~\ref{lemma}}
\begin{proof}
Given the original time series \( x_{t-L+1:t} \) with length \( L \), the downsampling process segments it into \( w \) subsequences, each of which contains every \( w \)-th data point from the original series. The length of each downsampled subsequence, denoted as \( n \), is therefore \( \left \lfloor \frac{L}{w} \right \rfloor \), as it collects one data point from every \( w \) time steps from the original series of length \( L \).

The SparseTSF model then applies a forecasting function \( f \) on each of these downsampled subsequences. The forecasting function \( f \) is designed to predict future values of the time series based on its past values. Specifically, it predicts the future subsequence \( x'_{t+1:t+m} \) using the past subsequence \( x'_{t-n+1:t} \). Here, \( m \) is the length of the forecast horizon for the downsampled subsequences and is given by \( \left \lfloor \frac{H}{w} \right \rfloor \), where \( H \) is the original forecast horizon.

Therefore, the SparseTSF model effectively reformulates the original forecasting task of predicting \( x_{t+1:t+H} \) from \( x_{t-L+1:t} \) into a series of smaller tasks. Each of these smaller tasks involves using the downsampled past subsequence \( x'_{t-n+1:t} \) to predict the downsampled future subsequence \( x'_{t+1:t+m} \). This is represented mathematically as:
\begin{align}
x'_{t+1:t+m} = f(x'_{t-n+1:t}).
\end{align}
\end{proof}

\paragraph{Proof of Theorem~\ref{theorem2}}
\begin{proof}
Theorem~\ref{theorem2} is established based on the assumption of a time series dataset that can be decomposed into a periodic component \( P(t) \) and a trend component \( T(t) \), as defined in Definition~\ref{definition}. This decomposition implies that any time point in the series \( X(t) \) can be expressed as the sum of its periodic and trend components, i.e., \( X(t) = P(t) + T(t) \).

Therefore, for the downsampled subsequences \(x'_{t-n+1:t}\) and \(x'_{t+1:t+m}\) based on a periodicity \( w \), we have:
\begin{align}
    x'_{t-n+1:t} = p'_{t-n+1:t}+t'_{t-n+1:t}, \\
    x'_{t+1:t+m} = p'_{t+1:t+m}+t'_{t+1:t+m}.
\end{align}

Hence, by combining with Lemma~\ref{lemma}, the task formulation of the SparseTSF model can be expressed as:
\begin{align}
p'_{t+1:t+m} + t'_{t+1:t+m} = f(p'_{t-n+1:t} + t'_{t-n+1:t}).
\end{align}

Due to the periodic nature of \( P(t) \) as defined in Equation~\ref{periodic_func}, for any two points \( i \) and \( j \) in the downsampled sequence (where \( i, j \in [t-n+1:t+m] \)), the periodic component remains constant, i.e., \( p'_i = p'_j \).

This indicates that the task of the SparseTSF model is to predict future trend components while utilizing a constant periodic component as a reference.
\end{proof}

\section{Case Study}
\label{case_study}

\subsection{Multi-Period Scenarios}
\label{multi_period_scenarios}

In this section, we specifically examine the performance of the SparseTSF model in scenarios involving multiple periods. Specifically, we study its performance on the Traffic dataset, as traffic flow data not only exhibits distinct daily periodicity but also demonstrates significant weekly cycles. For instance, the morning and evening rush hours represent intra-day cycles, while the different patterns between weekdays and weekends exemplify weekly cycles.

\begin{figure}[!htb]  
    \centering
    \subcaptionbox{Original}{\includegraphics[width=0.48\linewidth]{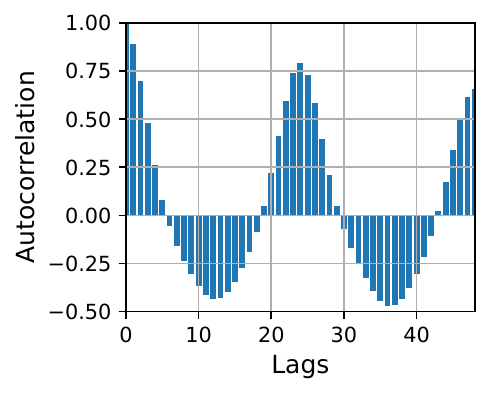}} 
    \subcaptionbox{Downsampled}{\includegraphics[width=0.48\linewidth]{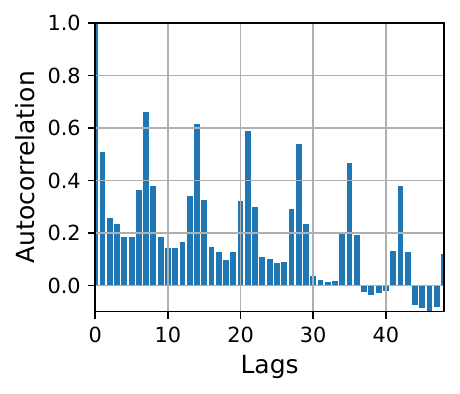}}
    \caption{Comparison of autocorrelation in original and downsampled subsequences for the last channel in the Traffic dataset.}
    \label{fig5}
\end{figure}

Figure~\ref{fig5} displays the autocorrelation in the original and day-period downsampled traffic flow data. It can be observed that even after downsampling with a daily period, the data still exhibits a clear weekly cycle (\(w'=7\)). Under these circumstances, with SparseTSF only decoupling the primary daily cycle, will it outperform the original fully connected linear model?

\begin{figure*}[!htb]  
    \centering
    \subcaptionbox{Linear}{\includegraphics[width=0.4\linewidth]{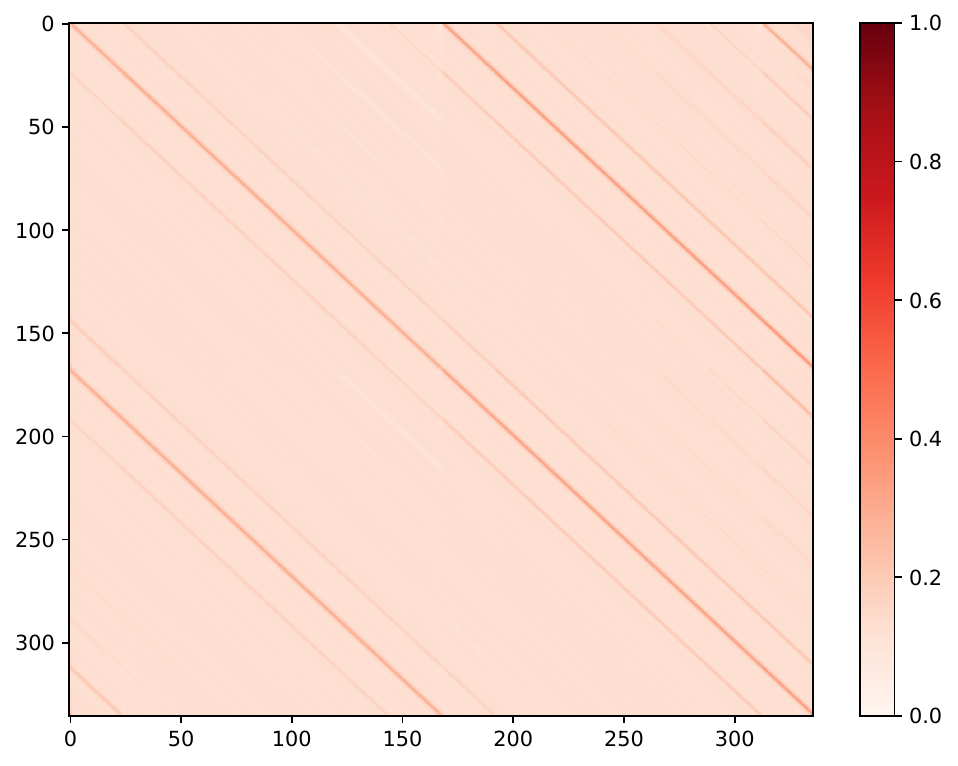}} 
    \subcaptionbox{SparseTSF}{\includegraphics[width=0.4\linewidth]{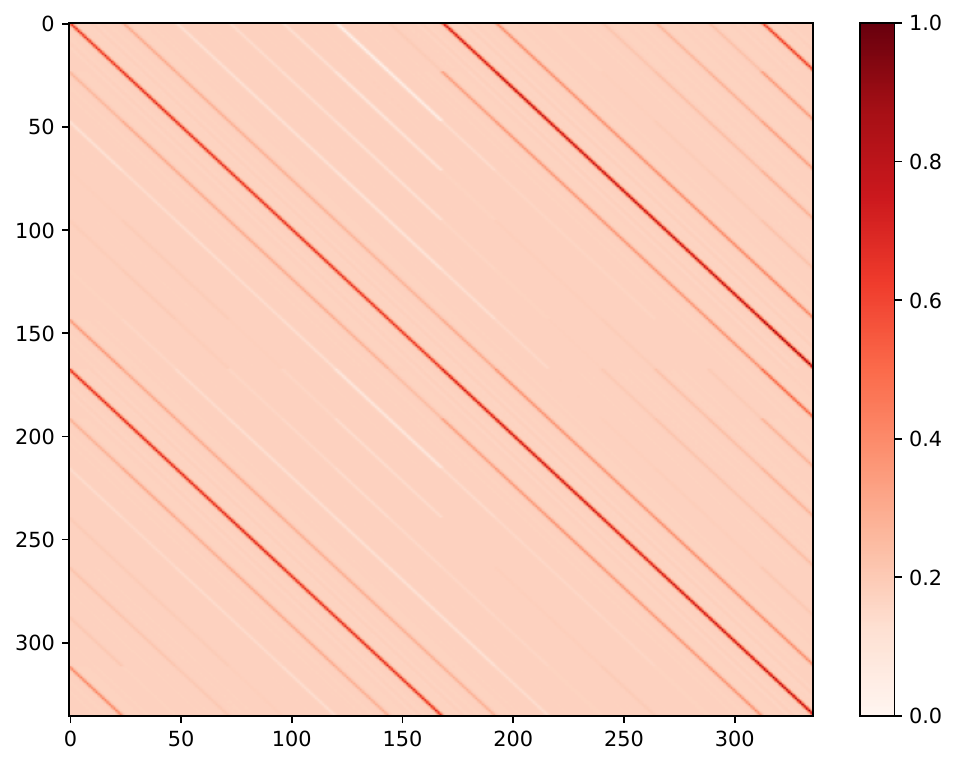}}
    \caption{Visualization of normalized weights of the model trained on the Traffic dataset with both look-back length (X-axis) and forecast horizon (Y-axis) of 336.}
    \label{fig6}
\end{figure*}

The results, as shown in Figure~\ref{fig6}, indicate that the SparseTSF model captures stronger daily and weekly periodic patterns (evident as more pronounced equidistant stripes) compared to the original approach. This is because, in the original method, a single linear layer is tasked with extracting both daily and weekly periodic patterns. In contrast, the SparseTSF model, by decoupling the daily cycle, simplifies the task for its inherent linear layer to only extract the remaining weekly periodic features. Therefore, even in scenarios with multiple periods, SparseTSF can still achieve remarkable performance.

\subsection{Ultra-Long Period Scenarios}
\label{long_period_scenarios}

This section is dedicated to examining the SparseTSF model's performance in scenarios characterized by ultra-long periods. Specifically, our focus is on the ETTm1\&ETTm2\footnote{https://github.com/zhouhaoyi/ETDataset} and Weather\footnote{https://www.bgc-jena.mpg.de/wetter} datasets, as detailed in Table~\ref{long_dataset}. These datasets are distinguished by their primary periods extending up to 96 and 144, respectively. We evaluate the SparseTSF model's performance under various settings of the hyperparameter \(w\).

\input{tables/long_dataset}

As illustrated in Table~\ref{long_period}, when \(w\) is set to a large value (for instance, 144, which aligns with the intrinsic primary period of the Weather dataset), the performance of the SparseTSF model tends to deteriorate. This decline is attributed to the excessive sparsity in connections caused by a large \(w\), limiting the information available for the model to base its predictions on, thereby impairing its performance. Interestingly, as \(w\) increases, there is a noticeable improvement in the SparseTSF model's performance. This observation suggests that employing denser connections within the SparseTSF framework could be a more viable option for datasets with longer periods.

Furthermore, an intriguing phenomenon is observed when \(w=1\), which corresponds to the scenario of employing a fully connected linear layer for prediction. The performance in this case is inferior compared to sparse connection-based predictions. This indicates that an appropriate level of sparsity in connections (even when the sparse interval does not match the dataset's inherent primary period) can enhance the model's predictive accuracy. This could be due to the redundant nature of time series data, especially when data sampling is dense. In such cases, executing sparse predictions might help eliminate some redundant information. However, these findings necessitate further investigation and exploration in future work.

\input{tables/long_period}

The findings above suggest that employing a denser sparse strategy would be beneficial in such cases. Therefore, we present in Table~\ref{long_result} a comparative performance of SparseTSF against other models under the setting of \(w=4\), where SparseTSF ranks within the top 3 in most cases. In this scenario, SparseTSF remains significantly lighter compared to other mainstream models. This indicates that the Sparse forecasting technique not only effectively reduces parameter size but also enhances prediction accuracy in most scenarios.

\input{tables/long_result}

\section{More Results and Analysis}
\label{more_result}

\subsection{Comparison Results after Fixing the Code Bug}
\label{fix_result}
\input{tables/main_wo_bug}

Recent research has discovered a long-standing bug in the popular codebase used in the field since the introduction of the Informer~\citep{informer}. This bug, which affected the calculation of test set metrics, caused the data that did not fill an entire batch to be discarded \citep{TFB}. As a result, the batch size setting influenced the results. Theoretically, the larger the batch size, the more test data might be discarded, leading to incorrect results. This bug significantly improved the performance on ETTh1 and ETTh2 datasets when the batch size was large, while the impact on other datasets was relatively minor.

To reassess the performance of SparseTSF, we present the performance of SparseTSF and existing models after fixing this bug in Table~\ref{main_wo_bug}. Here, we reran FITS under the conditions of lookback \(L=720\) and cutoff frequency \(COF=5\) (where the parameter count of SparseTSF is still tens of times smaller than that of FITS) for a fair comparison with SparseTSF. The results for other baselines were sourced from FITS' reproduction, where they reran the baselines' results after fixing the bug \citep{fits}. As shown, after fixing the code bug, SparseTSF still achieves impressive performance with minimal overhead, aligning with the conclusions of Table~\ref{main_reslut}.

\subsection{Impacts of Varying Look-Back Length}

The look-back length determines the richness of historical information the model can utilize. Generally, models are expected to perform better with longer input lengths if they possess robust long-term dependency modeling capabilities. Table~\ref{look_back} presents the performance of SparseTSF at different look-back lengths.

\input{tables/look_back}

It can be observed that two phenomena occur: (i) longer look-back windows perform better, indicating SparseTSF's ability in long-term dependency modeling, and (ii) the performance of the ETTh1 \& ETTh2 datasets remains relatively stable across different look-back windows, while the performance of the Traffic \& Electricity datasets varies significantly, especially with a look-back of 96, where the accuracy notably decreases.

In fact, we can further discuss the reasons behind the second point. As illustrated in Figure~\ref{fig3}, ETTh1 only exhibits a significant daily periodic pattern (\(w=24\)). In this case, look-back lengths of 96 can achieve good results because they fully encompass the daily periodic pattern. However, as shown in Figure~\ref{fig6}, Traffic not only has a significant daily periodic pattern (\(w=24\)) but also a noticeable weekly periodic pattern (\(w=168\)). In this case, a look-back of 96 cannot cover the entire weekly periodic pattern, leading to a significant performance drop. This underscores the necessity of sufficiently long look-back lengths (at least covering the entire cycle length) for accurate prediction. Given the extreme lightweight nature of SparseTSF, we strongly recommend providing sufficiently long look-back windows whenever feasible.

\subsection{Impacts of Instance Normalization}
Instance Normalization (IN) strategy has become popular in mainstream methods. We also employ this strategy in SparseTSF to enhance its performance on datasets with significant distribution drift. We showcase the impact of the IN strategy in Table~\ref{IN}.

\input{tables/in}

It can be observed that IN is necessary for smaller datasets, namely ETTh1 and ETTh2 datasets. However, its effect is relatively limited on larger datasets such as Traffic and Electricity datasets. It must be clarified that, although the IN strategy is one of the factors contributing to SparseTSF's success, it is not the key differentiator of SparseTSF's core contributions compared to other models.

\subsection{Comparison Results with N-HiTS and OneShotSTL}
\label{comp_other}
\input{tables/comp_other}

Here, we present the comparison results between SparseTSF and N-HiTS and OneShotSTL in Table~\ref{comp_other_table}. It can be observed that in most cases, SparseTSF outperforms these methods, demonstrating the superiority of the SparseTSF approach.


\end{document}

%% file: tables/dataset.tex
\begin{table}[htb]
\centering
\caption{Summary of datasets.}
\begin{adjustbox}{max width=0.75\columnwidth}
\begin{tabular}{@{}lccccccc@{}}
\toprule
Datasets & ETTh1 \& ETTh2 & Electricity & Traffic \\ \midrule
Channels & 7 & 321 & 862 \\
Frequency & hourly & hourly & hourly \\
Timesteps & 17,420 & 26,304 & 17,544 \\ \bottomrule
\end{tabular}
\end{adjustbox}
\label{dataset}
\end{table}

%% file: tables/main_result.tex
\begin{table*}[!htb]
\centering
\caption{MSE results of multivariate long-term time series forecasting comparing SparseTSF with other mainstream models. The top two results are highlighted in \textbf{bold}. The reported results of SparseTSF are \textit{averaged} over 5 runs with \textit{standard deviation} included. 'Imp.' denotes the improvement compared to the best-performing baseline models.}
\label{main_reslut}
\begin{adjustbox}{max width=\linewidth}
\begin{tabular}{@{}c|cccc|cccc|cccc|cccc@{}}
\toprule
Dataset & \multicolumn{4}{c|}{ETTh1} & \multicolumn{4}{c|}{ETTh2} & \multicolumn{4}{c|}{Electricity} & \multicolumn{4}{c}{Traffic} \\ \midrule
Horizon & 96 & 192 & 336 & 720 & 96 & 192 & 336 & 720 & 96 & 192 & 336 & 720 & 96 & 192 & 336 & 720 \\ \midrule
Informer~\citeyearpar{informer} & 0.865 & 1.008 & 1.107 & 1.181 & 3.755 & 5.602 & 4.721 & 3.647 & 0.274 & 0.296 & 0.300 & 0.373 & 0.719 & 0.696 & 0.777 & 0.864 \\
Autoformer~\citeyearpar{autoformer} & 0.449 & 0.500 & 0.521 & 0.514 & 0.358 & 0.456 & 0.482 & 0.515 & 0.201 & 0.222 & 0.231 & 0.254 & 0.613 & 0.616 & 0.622 & 0.660 \\
Pyraformer~\citeyearpar{pyraformer} & 0.664 & 0.790 & 0.891 & 0.963 & 0.645 & 0.788 & 0.907 & 0.963 & 0.386 & 0.386 & 0.378 & 0.376 & 2.085 & 0.867 & 0.869 & 0.881 \\
FEDformer~\citeyearpar{fedformer} & 0.376 & 0.420 & 0.459 & 0.506 & 0.346 & 0.429 & 0.496 & 0.463 & 0.193 & 0.201 & 0.214 & 0.246 & 0.587 & 0.604 & 0.621 & 0.626 \\
FiLM~\citeyearpar{film} & 0.371 & 0.414 & 0.442 & 0.465 & 0.284 & 0.357 & 0.377 & 0.439 & 0.154 & 0.164 & 0.188 & 0.236 & 0.416 & 0.408 & 0.425 & 0.520 \\
TimesNet~\citeyearpar{timesnet} & 0.384 & 0.436 & 0.491 & 0.521 & 0.340 & 0.402 & 0.452 & 0.462 & 0.168 & 0.184 & 0.198 & 0.220 & 0.593 & 0.617 & 0.629 & 0.640 \\
PatchTST~\citeyearpar{patchtst} & \textbf{0.370} & 0.413 & \textbf{0.422} & 0.447 & \textbf{0.274} & 0.341 & \textbf{0.329} & 0.379 & \textbf{0.129} & \textbf{0.147} & \textbf{0.163} & \textbf{0.197} & \textbf{0.360} & \textbf{0.379} & \textbf{0.392} & \textbf{0.432} \\
DLinear~\citeyearpar{dlinear} & 0.374 & \textbf{0.405} & 0.429 & 0.440 & 0.338 & 0.381 & 0.400 & 0.436 & 0.140 & 0.153 & 0.169 & \textbf{0.203} & 0.410 & 0.423 & 0.435 & 0.464 \\
FITS~\citeyearpar{fits} & 0.375 & 0.408 & 0.429 & \textbf{0.427} & \textbf{0.274} & \textbf{0.333} & 0.340 & \textbf{0.374} & \textbf{0.138} & 0.152 & 0.166 & 0.205 & 0.401 & 0.407 & 0.420 & 0.456 \\ \midrule
\multirow{2}{*}{SparseTSF (ours)} & \textbf{0.359} & \textbf{0.397} & \textbf{0.404} & \textbf{0.417} & \textbf{0.267} & \textbf{0.314} & \textbf{0.312} & \textbf{0.370} & \textbf{0.138} & \textbf{0.146} & \textbf{0.164} & \textbf{0.203} & \textbf{0.382} & \textbf{0.388} & \textbf{0.402} & \textbf{0.445} \\
 & $\pm$0.006 & $\pm$0.002 & $\pm$0.001 & $\pm$0.001 & $\pm$0.005 & $\pm$0.003 & $\pm$0.004 & $\pm$0.001 & $\pm$0.001 & $\pm$0.001 & $\pm$0.001 & $\pm$0.001 & $\pm$0.001 & $\pm$0.001 & $\pm$0.001 & $\pm$0.002 \\ \midrule
Imp. & +0.011 & +0.008 & +0.018 & +0.010 & +0.007 & +0.019 & +0.017 & +0.004 & -0.009 & +0.001 & -0.001 & -0.006 & -0.022 & -0.009 & -0.010 & -0.013 \\ \bottomrule
\end{tabular}
\end{adjustbox}
\end{table*}

%% file: tables/measure.tex
\begin{table}[!htb]
\centering
\caption{Static and runtime metrics of SparseTSF and other mainstream models on the Electricity Dataset with a forecast horizon of 720. Here, the look-back length for each model is set to be consistent with their respective official papers, such as 336 for DLinear and 720 for FITS.}
\label{measure}
\begin{adjustbox}{max width=\linewidth}
\begin{tabular}{@{}c|cccc@{}}
\toprule
Model & Parameters & MACs & Max Mem.(MB) & Epoch Time(s) \\ \midrule
Informer~\citeyearpar{informer} & 12.53 M & 3.97 G & 969.7 & 70.1 \\
Autoformer~\citeyearpar{autoformer} & 12.22 M & 4.41 G & 2631.2 & 107.7 \\
FEDformer~\citeyearpar{fedformer} & 17.98 M & 4.41 G & 1102.5 & 238.7 \\
FiLM~\citeyearpar{film} & 12.22 M & 4.41 G & 1773.9 & 78.3 \\
PatchTST~\citeyearpar{patchtst} & 6.31 M & 11.21 G & 10882.3 & 290.3 \\
\midrule
DLinear~\citeyearpar{dlinear} & 485.3 K & 156.0 M & 123.8 & 25.4 \\
FITS~\citeyearpar{fits} & 10.5 K & 79.9 M & 496.7 & 35.0 \\
SparseTSF (Ours) & \textbf{0.92 K} & \textbf{12.71 M} & 125.2 & 31.3 \\ \bottomrule
\end{tabular}
\end{adjustbox}
\end{table}

%% file: tables/parameters.tex
\begin{table}[!htb]
\centering
\caption{Comparison of the scale of parameters on Electricity dataset between SparseTSF and FITS models under different configurations of look-back length and forecast horizon, where SparseTSF operates with \(w=24\) and FITS employs COF at the \(2^{th}\) harmonic.}
\label{parameters}
\begin{adjustbox}{max width=\linewidth}
\begin{tabular}{c|cccc|cccc}
\toprule
Model & \multicolumn{4}{c|}{SparseTSF (Ours)} & \multicolumn{4}{c}{FITS~\citeyearpar{fits}} \\ \midrule
\diagbox{Horizon}{Look-back} & 96 & 192 & 336 & 720 & 96 & 192 & 336 & 720 \\ \midrule
\multicolumn{1}{c|}{96} & 41 & 57 & 81 & \multicolumn{1}{c|}{145} & 840 & 1,218 & 2,091 & 5,913 \\
\multicolumn{1}{c|}{192} & 57 & 89 & 137 & \multicolumn{1}{c|}{265} & 1,260 & 1,624 & 2,542 & 6,643 \\
\multicolumn{1}{c|}{336} & 81 & 137 & 221 & \multicolumn{1}{c|}{445} & 1,890 & 2,233 & 3,280 & 7,665 \\
\multicolumn{1}{c|}{720} & 145 & 265 & 445 & \multicolumn{1}{c|}{925} & 3,570 & 3,857 & 5,125 & 10,512 \\ \bottomrule
\end{tabular}
\end{adjustbox}
\end{table}

%% file: tables/ablation.tex
\begin{table}[!htb]
\centering
\caption{Ablation MSE results of the Sparse technique. All results are collected with a unified channel-independent and instance normalization strategy. The 'Boost' indicates the percentage of performance improvement after incorporating the Sparse technique.}
\label{ablation}
\begin{adjustbox}{max width=\linewidth}
\begin{tabular}{@{}c|cccc|cccc@{}}
\toprule
Dataset & \multicolumn{4}{c|}{ETTh1} & \multicolumn{4}{c}{ETTh2} \\ \midrule
Horizon & 96 & 192 & 336 & 720 & 96 & 192 & 336 & 720 \\ \midrule
Linear & 0.371 & 0.460 & 0.417 & 0.424 & \textbf{0.257} & 0.337 & 0.336 & 0.391 \\
+sparse & \textbf{0.359} & \textbf{0.397} & \textbf{0.404} & \textbf{0.417} & 0.267 & \textbf{0.314} & \textbf{0.312} & \textbf{0.370} \\
Boost & 3.3\% & 13.8\% & 3.1\% & 1.7\% & -3.9\% & 6.9\% & 7.1\% & 5.3\% \\ \midrule
Transformer & 0.697 & 0.732 & 0.714 & 0.770 & 0.340 & \textbf{0.376} & 0.366 & 0.468 \\
+sparse & \textbf{0.406} & \textbf{0.442} & \textbf{0.446} & \textbf{0.489} & \textbf{0.322} & 0.380 & \textbf{0.353} & \textbf{0.432} \\
Boost & 41.7\% & 39.6\% & 37.5\% & 36.5\% & 5.2\% & -1.0\% & 3.6\% & 7.7\% \\ \midrule
GRU & 0.415 & 0.529 & 0.512 & 0.620 & 0.296 & 0.345 & 0.363 & 0.454 \\
+sparse & \textbf{0.356} & \textbf{0.391} & \textbf{0.437} & \textbf{0.455} & \textbf{0.282} & \textbf{0.332} & \textbf{0.356} & \textbf{0.421} \\
Boost & 14.1\% & 26.1\% & 14.7\% & 26.7\% & 4.8\% & 3.7\% & 1.9\% & 7.2\% \\ \bottomrule
\end{tabular}
\end{adjustbox}
\end{table}

%% file: tables/period.tex
\begin{table}[!htb]
\centering
\caption{MSE results of SparseTSF on ETTh1 with varied hyperparameters \(w\).}
\label{period}
\begin{adjustbox}{max width=\linewidth}
\begin{tabular}{c|cccc|ccc}
\toprule
Horizon & \begin{tabular}[c]{@{}c@{}}SparseTSF\\ (\(w\)=6)\end{tabular} & \begin{tabular}[c]{@{}c@{}}SparseTSF\\ (\(w\)=12)\end{tabular} & \begin{tabular}[c]{@{}c@{}}SparseTSF\\ (\(w\)=24)\end{tabular} & \begin{tabular}[c]{@{}c@{}}SparseTSF\\ (\(w\)=48)\end{tabular} & \begin{tabular}[c]{@{}c@{}}FITS\\ \citeyearpar{fits} \end{tabular} & \begin{tabular}[c]{@{}c@{}}DLinear\\ \citeyearpar{dlinear} \end{tabular} & \begin{tabular}[c]{@{}c@{}}PatchTST\\ \citeyearpar{patchtst} \end{tabular} \\ \midrule
96 & 0.376 & 0.369 & \textbf{0.359} & 0.380 & 0.375 & 0.374 & 0.370 \\
192 & 0.410 & 0.402 & \textbf{0.397} & 0.400 & 0.408 & 0.405 & 0.413 \\
336 & 0.408 & 0.406 & 0.404 & \textbf{0.399} & 0.429 & 0.429 & 0.422 \\
720 & 0.427 & 0.423 & \textbf{0.417} & 0.427 & 0.427 & 0.440 & 0.447 \\\midrule
Avg. & 0.405 & 0.400 & \textbf{0.394} & 0.402 & 0.410 & 0.412 & 0.413 \\ \bottomrule
\end{tabular}
\end{adjustbox}
\end{table}

%% file: tables/cross_domain.tex
\begin{table}[!htb]
\centering
\caption{Comparison of generalization capabilities between SparseTSF and other mainstream models. 'Dataset A \(\to\) Dataset B' indicates training and validation on the training and validation sets of Dataset A, followed by testing on the test set of Dataset B.}
\label{cross_domain}
\begin{adjustbox}{max width=\linewidth}
\begin{tabular}{@{}c|cccc|cccc@{}}
\toprule
Dataset & \multicolumn{4}{c|}{ETTh2 \(\to\) ETTh1} & \multicolumn{4}{c}{Electricity \(\to\) ETTh1} \\ \midrule
Horizon & 96 & 192 & 336 & 720 & 96 & 192 & 336 & 720 \\ \midrule
\multicolumn{1}{c|}{Informer~\citeyearpar{informer}} & 0.844 & 0.921 & 0.898 & \multicolumn{1}{c|}{0.829} & \textbackslash{} & \textbackslash{} & \textbackslash{} & \textbackslash{} \\
\multicolumn{1}{c|}{Autoformer~\citeyearpar{autoformer}} & 0.978 & 1.058 & 0.944 & \multicolumn{1}{c|}{0.921} & \textbackslash{} & \textbackslash{} & \textbackslash{} & \textbackslash{} \\
\multicolumn{1}{c|}{FEDformer~\citeyearpar{fedformer}} & 0.878 & 0.927 & 0.939 & \multicolumn{1}{c|}{0.967} & \textbackslash{} & \textbackslash{} & \textbackslash{} & \textbackslash{} \\
\multicolumn{1}{c|}{FiLM~\citeyearpar{film}} & 0.876 & 0.904 & 0.919 & \multicolumn{1}{c|}{0.925} & \textbackslash{} & \textbackslash{} & \textbackslash{} & \textbackslash{} \\
\multicolumn{1}{c|}{PatchTST~\citeyearpar{patchtst}} & 0.449 & 0.478 & 0.482 & \multicolumn{1}{c|}{0.476} & 0.400 & 0.424 & 0.475 & 0.472 \\
\multicolumn{1}{c|}{DLinear~\citeyearpar{dlinear}} & 0.430 & 0.478 & 0.458 & \multicolumn{1}{c|}{0.506} & 0.397 & 0.428 & 0.447 & 0.470 \\
\multicolumn{1}{c|}{Fits~\citeyearpar{fits}} & 0.419 & 0.427 & 0.428 & \multicolumn{1}{c|}{0.445} & 0.380 & 0.414 & 0.440 & 0.448 \\
\multicolumn{1}{c|}{SparseTSF (Ours)} & \textbf{0.370} & \textbf{0.401} & \textbf{0.412} & \multicolumn{1}{c|}{\textbf{0.419}} & \textbf{0.373} & \textbf{0.409} & \textbf{0.433} & \textbf{0.439} \\ \bottomrule
\end{tabular}
\end{adjustbox}
\end{table}

%% file: tables/long_dataset.tex
\begin{table}[htb]
\centering
\caption{Summary of datasets with ultra-long periods.}
\begin{adjustbox}{max width=0.65\columnwidth}
\begin{tabular}{@{}lccccccc@{}}
\toprule
Datasets & ETTm1 & ETTm2 & Weather \\ \midrule
Channels & 7 & 7 & 21 \\
Frequency & 15 mins & 15 mins & 10 mins \\
Timesteps &  69,680 &  69,680 & 52,696 \\ \bottomrule
\end{tabular}
\end{adjustbox}
\label{long_dataset}
\end{table}

%% file: tables/long_period.tex
\begin{table}[!htb]
\centering
\caption{MSE results of SparseTSF on ultra-long period datasets with varied hyperparameters \(w\). The forecast horizon is set as 720.}
\label{long_period}
\begin{adjustbox}{max width=\linewidth}
\begin{tabular}{@{}c|cccccccc@{}}
\toprule
\multirow{2}{*}{Dataset} & \multicolumn{8}{c}{Parameter w} \\ \cmidrule(l){2-9} 
 & 144 & 72 & 48 & 24 & 12 & 6 & 2 & 1 \\ \midrule
ETTm1 & 0.450 & 0.450 & 0.422 & 0.422 & 0.421 & \textbf{0.415} & \textbf{0.415} & 0.429 \\
ETTm2 & 0.375 & 0.371 & 0.373 & 0.352 & 0.354 & \textbf{0.349} & \textbf{0.349} & 0.357 \\
Weather & 0.332 & 0.329 & 0.325 & 0.321 & 0.319 & 0.319 & \textbf{0.318} & 0.322 \\ \bottomrule
\end{tabular}
\end{adjustbox}
\end{table}

%% file: tables/long_result.tex
\begin{table*}[!htb]
\centering
\caption{MSE results on ultra-long period datasets comparing SparseTSF (\(w=4\)) with other mainstream models. The ranking of SparseTSF's performance is shown in parentheses.}
\label{long_result}
\begin{adjustbox}{max width=0.9\linewidth}
\begin{tabular}{@{}c|cccc|cccc|cccc@{}}
\toprule
Dataset & \multicolumn{4}{c|}{ETTm1} & \multicolumn{4}{c|}{ETTm2} & \multicolumn{4}{c}{Weather} \\ \midrule
Horizon & \multicolumn{1}{c}{96} & \multicolumn{1}{c}{192} & \multicolumn{1}{c}{336} & \multicolumn{1}{c|}{720} & \multicolumn{1}{c}{96} & \multicolumn{1}{c}{192} & \multicolumn{1}{c}{336} & \multicolumn{1}{c|}{720} & \multicolumn{1}{c}{96} & \multicolumn{1}{c}{192} & \multicolumn{1}{c}{336} & \multicolumn{1}{c}{720} \\ \midrule
Informer~\citeyearpar{informer} & 0.672 & 0.795 & 1.212 & 1.166 & 0.365 & 0.533 & 1.363 & 3.379 & 0.300 & 0.598 & 0.578 & 1.059 \\
Autoformer~\citeyearpar{autoformer} & 0.505 & 0.553 & 0.621 & 0.671 & 0.255 & 0.281 & 0.339 & 0.433 & 0.266 & 0.307 & 0.359 & 0.419 \\
Pyraformer~\citeyearpar{pyraformer} & 0.543 & 0.557 & 0.754 & 0.908 & 0.435 & 0.730 & 1.201 & 3.625 & 0.896 & 0.622 & 0.739 & 1.004 \\
FEDformer~\citeyearpar{fedformer} & 0.379 & 0.426 & 0.445 & 0.543 & 0.203 & 0.269 & 0.325 & 0.421 & 0.217 & 0.276 & 0.339 & 0.403 \\
TimesNet~\citeyearpar{timesnet} & 0.338 & 0.374 & 0.410 & 0.478 & 0.187 & 0.249 & 0.321 & 0.408 & 0.172 & 0.219 & 0.280 & 0.365 \\
PatchTST~\citeyearpar{patchtst} & 0.293 & 0.333 & 0.369 & 0.416 & 0.166 & 0.223 & 0.274 & 0.362 & 0.149 & 0.194 & 0.245 & 0.314 \\
DLinear~\citeyearpar{dlinear} & 0.299 & 0.335 & 0.369 & 0.425 & 0.167 & 0.221 & 0.274 & 0.368 & 0.176 & 0.218 & 0.262 & 0.323 \\
FITS~\citeyearpar{fits} & 0.305 & 0.339 & 0.367 & 0.418 & 0.164 & 0.217 & 0.269 & 0.347 & 0.145 & 0.188 & 0.236 & 0.308 \\ \midrule
SparseTSF (ours) & 0.314(4) & 0.343(4) & 0.369(2) & 0.418(2) & 0.165(2) & 0.218(2) & 0.272(2) & 0.35(2) & 0.172(3) & 0.215(3) & 0.26(3) & 0.318(3) \\ \bottomrule
\end{tabular}
\end{adjustbox}
\end{table*}

%% file: tables/main_wo_bug.tex
\begin{table*}[!htb]
\centering
\caption{MSE results of multivariate long-term time series forecasting comparing SparseTSF with other mainstream models after fixing code bug. The top two results are highlighted in \textbf{bold}.}
\label{main_wo_bug}
\begin{adjustbox}{max width=\linewidth}
\begin{tabular}{@{}c|cccc|cccc|cccc|cccc@{}}
\toprule
Dataset & \multicolumn{4}{c|}{ETTh1} & \multicolumn{4}{c|}{ETTh2} & \multicolumn{4}{c|}{Electricity} & \multicolumn{4}{c}{Traffic} \\ \midrule
Horizon & 96 & 192 & 336 & 720 & 96 & 192 & 336 & 720 & 96 & 192 & 336 & 720 & 96 & 192 & 336 & 720 \\ \midrule
FEDformer~\citeyearpar{fedformer} & \textbf{0.375} & 0.427 & 0.459 & 0.484 & 0.340 & 0.433 & 0.508 & 0.480 & 0.188 & 0.197 & 0.212 & 0.244 & 0.573 & 0.611 & 0.621 & 0.630 \\
TimesNet~\citeyearpar{timesnet} & 0.384 & 0.436 & 0.491 & 0.521 & 0.340 & 0.402 & 0.452 & 0.462 & 0.168 & 0.184 & 0.198 & 0.220 & 0.593 & 0.617 & 0.629 & 0.640 \\
PatchTST~\citeyearpar{patchtst} & 0.385 & \textbf{0.413} & 0.440 & 0.456 & \textbf{0.274} & \textbf{0.338} & 0.367 & 0.391 & \textbf{0.129} & \textbf{0.149} & \textbf{0.166} & 0.210 & \textbf{0.366} & \textbf{0.388} & \textbf{0.398} & \textbf{0.457} \\
DLinear~\citeyearpar{dlinear} & 0.384 & 0.443 & 0.446 & 0.504 & 0.282 & 0.350 & 0.414 & 0.588 & 0.140 & 0.153 & 0.169 & \textbf{0.204} & 0.413 & 0.423 & 0.437 & 0.466 \\
FITS~\citeyearpar{fits} & 0.382 & 0.417 & \textbf{0.436} & \textbf{0.433} & \textbf{0.272} & \textbf{0.333} & \textbf{0.355} & \textbf{0.378} & 0.145 & 0.159 & 0.175 & 0.212 & 0.398 & 0.409 & 0.421 & \textbf{0.457} \\
SparseTSF (ours) & \textbf{0.362} & \textbf{0.403} & \textbf{0.434} & \textbf{0.426} & 0.294 & 0.339 & \textbf{0.359} & \textbf{0.383} & \textbf{0.138} & \textbf{0.151} & \textbf{0.166} & \textbf{0.205} & \textbf{0.389} & \textbf{0.398} & \textbf{0.411} & \textbf{0.448} \\ \bottomrule
\end{tabular}
\end{adjustbox}
\end{table*}

%% file: tables/look_back.tex
\begin{table*}[!htb]
\centering
\caption{MSE results of SparseTSF with varied look-back lengths.}
\label{look_back}
\begin{adjustbox}{max width=0.9\linewidth}
\begin{tabular}{c|cccc|cccc|cccc|cccc}
\toprule
Dataset & \multicolumn{4}{c|}{ETTh1} & \multicolumn{4}{c|}{ETTh2} & \multicolumn{4}{c|}{Electricity} & \multicolumn{4}{c}{Traffic} \\ \midrule
\diagbox{Horizon}{Look-back} & 96 & 192 & 336 & 720 & 96 & 192 & 336 & 720 & 96 & 192 & 336 & 720 & 96 & 192 & 336 & 720 \\ \midrule
96 & 0.380 & 0.371 & 0.393 & 0.354 & 0.288 & 0.285 & 0.272 & 0.278 & 0.209 & 0.160 & 0.146 & 0.138 & 0.672 & 0.455 & 0.412 & 0.383 \\
192 & 0.433 & 0.434 & 0.418 & 0.398 & 0.363 & 0.346 & 0.323 & 0.315 & 0.202 & 0.166 & 0.154 & 0.147 & 0.608 & 0.453 & 0.415 & 0.388 \\
336 & 0.447 & 0.420 & 0.390 & 0.405 & 0.366 & 0.335 & 0.314 & 0.311 & 0.217 & 0.184 & 0.172 & 0.164 & 0.609 & 0.468 & 0.428 & 0.403 \\
720 & 0.451 & 0.426 & 0.413 & 0.418 & 0.407 & 0.389 & 0.372 & 0.371 & 0.259 & 0.223 & 0.210 & 0.205 & 0.650 & 0.493 & 0.462 & 0.446 \\ \midrule
\textbf{Avg.} & 0.428 & 0.413 & 0.404 & \textbf{0.394} & 0.356 & 0.339 & 0.320 & \textbf{0.319} & 0.222 & 0.183 & 0.171 & \textbf{0.163} & 0.635 & 0.467 & 0.429 & \textbf{0.405} \\ \bottomrule
\end{tabular}
\end{adjustbox}
\end{table*}

%% file: tables/in.tex
\begin{table}[!htb]
\centering
\caption{Ablation results of IN strategy in SparseTSF.}
\label{IN}
\begin{adjustbox}{max width=\linewidth}
\begin{tabular}{@{}c|cc|cc|cc|cc@{}}
\toprule
Dataset & \multicolumn{2}{c|}{ETTh1} & \multicolumn{2}{c|}{ETTh2} & \multicolumn{2}{c|}{Electricity} & \multicolumn{2}{c}{Traffic} \\ \midrule
Horizon & w/ IN & w/o IN & w/ IN & w/o IN & w/ IN & w/o IN & w/ IN & w/o IN \\ \midrule
96 & \textbf{0.359} & 0.37 & \textbf{0.267} & 0.327 & \textbf{0.138} & \textbf{0.138} & \textbf{0.382} & \textbf{0.382} \\
192 & \textbf{0.397} & 0.413 & \textbf{0.314} & 0.426 & \textbf{0.146} & \textbf{0.146} & 0.388 & \textbf{0.387} \\
336 & \textbf{0.404} & 0.431 & \textbf{0.312} & 0.482 & 0.164 & \textbf{0.163} & 0.402 & \textbf{0.401} \\
720 & \textbf{0.417} & 0.462 & \textbf{0.37} & 0.866 & 0.203 & \textbf{0.198} & 0.445 & \textbf{0.444} \\ \bottomrule
\end{tabular}
\end{adjustbox}
\end{table}

%% file: tables/comp_other.tex
\begin{table}[!htb]
\centering
\caption{Comparison Results with N-HiTS and OneShotSTL. In this comparison, SparseTSF and N-HiTS are evaluated based on multivariate prediction results (MSE), while SparseTSF and OneShotSTL are compared using univariate prediction results (MAE). Their results are sourced from their respective official papers.}
\label{comp_other_table}
\begin{adjustbox}{max width=0.95\linewidth}
\begin{tabular}{@{}c|c|cc|cc@{}}
\toprule
Dataset & Horizon & Nhit & SparseTSF & OneShotSTL & SparseTSF \\ \midrule
\multirow{4}{*}{ETTm2} & 96 & 0.176 & \textbf{0.165} & 0.211 & \textbf{0.187} \\
 & 192 & 0.245 & \textbf{0.218} & 0.244 & \textbf{0.233} \\
 & 336 & 0.295 & \textbf{0.272} & 0.273 & \textbf{0.268} \\
 & 720 & 0.401 & \textbf{0.350} & \textbf{0.321} & 0.324 \\ \midrule
\multirow{4}{*}{Electricity} & 96 & 0.147 & \textbf{0.138} & 0.331 & \textbf{0.314} \\
 & 192 & 0.167 & \textbf{0.146} & 0.355 & \textbf{0.334} \\
 & 336 & 0.186 & \textbf{0.164} & 0.389 & \textbf{0.366} \\
 & 720 & 0.243 & \textbf{0.203} & 0.444 & \textbf{0.416} \\ \midrule
\multirow{4}{*}{Traffic} & 96 & 0.402 & \textbf{0.382} & 0.181 & \textbf{0.179} \\
 & 192 & 0.42 & \textbf{0.388} & 0.181 & \textbf{0.175} \\
 & 336 & 0.448 & \textbf{0.402} & \textbf{0.182} & 0.184 \\
 & 720 & 0.539 & \textbf{0.445} & \textbf{0.199} & 0.203 \\ \bottomrule
\end{tabular}
\end{adjustbox}
\end{table}